\pdfoutput=1

\documentclass[11pt]{article}

\usepackage[]{acl}

\usepackage{ulem}
\usepackage{times}
\usepackage{latexsym}
\usepackage{multirow}
\usepackage{graphicx}
\usepackage{enumitem}
\usepackage[linesnumbered,ruled,vlined]{algorithm2e} 
\usepackage{amsmath}
\usepackage{amssymb}
\usepackage{latexsym}
\usepackage[inkscapelatex=false]{svg}
\usepackage{url}

\SetKwInput{KwInput}{Input}                
\SetKwInput{KwOutput}{Output}              
\usepackage{ulem}   

\usepackage{makecell}
\usepackage{color}
\usepackage{booktabs}
\usepackage{hyperref}
\usepackage[all]{nowidow}

\usepackage[T1]{fontenc}

\usepackage[utf8]{inputenc}

\usepackage{microtype}

\title{NGEP: A Graph-based Event Planning Framework for Story Generation}

\author{Chen Tang\textsuperscript{1}, Zhihao Zhang\textsuperscript{2}, Tyler Loakman\textsuperscript{3}, Chenghua Lin\textsuperscript{3}\footnotemark[1] ~and Frank Guerin\textsuperscript{1} \\
\textsuperscript{1}Department of Computer Science, The University of Surrey, UK \\
\textsuperscript{2}School of Economics and Management, Beihang University, Beijing, China \\
\textsuperscript{3}Department of Computer Science, The University of Sheffield, UK \\
\texttt{\{chen.tang,f.guerin\}@surrey.ac.uk} \\ \texttt{zhhzhang@buaa.edu.cn} \\
\texttt{\{tcloakman1,c.lin\}@sheffield.ac.uk}}

\begin{document}
\maketitle

\renewcommand{\thefootnote}{\fnsymbol{footnote}} 
\footnotetext[1]{Corresponding author.} 
\renewcommand{\thefootnote}{\arabic{footnote}} 

\begin{abstract}
To improve the performance of long text generation, recent studies have leveraged automatically planned event structures (i.e. storylines) to guide story generation. Such prior works mostly employ end-to-end neural generation models to predict event sequences for a story. However, such generation models struggle to guarantee the narrative coherence of separate events due to the hallucination problem, and additionally the generated event sequences are often hard to control due to the end-to-end nature of the models. To address these challenges, we propose NGEP, an novel event planning framework which generates an event sequence by performing inference on an automatically constructed event graph and enhances generalisation ability through a neural event advisor. We conduct a range of experiments on multiple criteria, and the results demonstrate that our graph-based neural framework outperforms the state-of-the-art (SOTA) event planning approaches, considering both the performance of event sequence generation and the effectiveness on the downstream task of story generation.
\end{abstract}

\section{Introduction}
Current neural generation models struggle to generate long stories as it is  difficult to guarantee the logical coherence of generated sentences when conditioning only on a limited size input (e.g. leading context or title). Therefore, current story generation frameworks are usually split into two stages, planning and writing, using an automatically planned storyline (aka. event sequence) \cite{alhussain2021automatic, tang2022recent} as the intermediate between planning and writing. 

In order to plan an event sequence, prior works \cite{martin2018event, yao2019plan, chen-etal-2021-graphplan, alhussain2021automatic, wang-etal-2020-plan} mostly focus on leveraging end-to-end neural generation models, such as BART \cite{lewis-etal-2020-bart}, to predict events. However, whilst some efforts \cite{goldfarb-tarrant-etal-2020-content, ahn-etal-2016-improving} have been made to improve neural event planning (e.g., \citet{goldfarb-tarrant-etal-2020-content} use rescoring models to guide the planning process), event planning based on neural generation models still tends to suffer from common limitations: (i) The selection of individual events in the sequence is hard to control (because of the end-to-end generation) \cite{chen-etal-2021-graphplan}; and (ii) Due to the hallucination problem \cite{rohrbach-etal-2018-object, elder-etal-2020-make, cheng-etal-2021-guiding, tang2022recent} each predicted event is not guaranteed to be complete and accurate.

\begin{figure}[tb]
\centering
\includegraphics[width=\columnwidth]{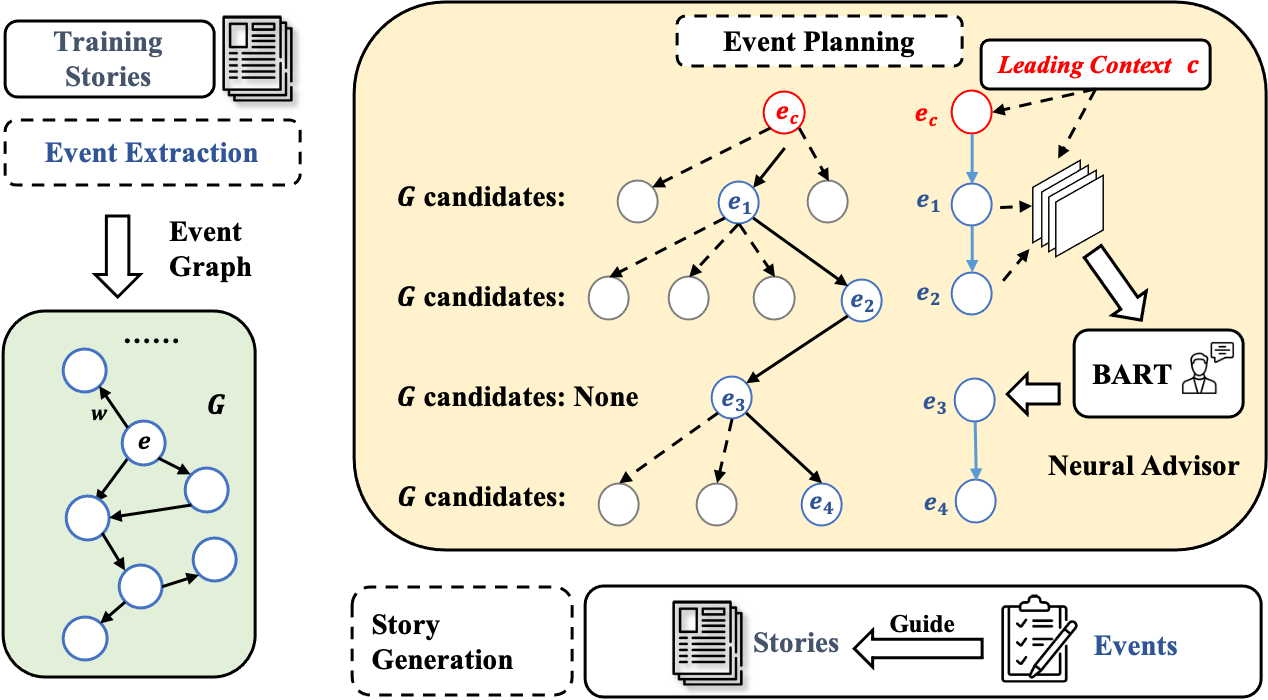}
\caption{The overview of our proposed NGEP model. The event graph $G$ is automatically constructed from the training set, and the potential event candidates are generated according to the conditional probability distribution modelled on $G$ when event planning. If there are no proper candidates for the next event, we leverage a BART-based neural advisor to predict the best choice.}
\label{fig:overview}
\end{figure}


In this study, we propose \textbf{NGEP}, a novel \textbf{N}eural \textbf{G}raph-based \textbf{E}vent \textbf{P}lanning framework to predict event sequences for story generation. An overview of the proposed framework is illustrated in \autoref{fig:overview}. Firstly, events are extracted from the training set in order to construct an event graph which records the events and their neighbour relations. This graph can then be used at test-time to predict events from a leading context. The conditional probability distribution is modelled by a coherence score calculated with the degrees of event nodes and the concurrency of predicted events. When an event graph is unable to generate event candidates, i.e. no edges point to another event, a BART-based neural advisor is introduced to predict the next event from the graph. The neural advisor is trained to model the conditional probability between event nodes and the context, including the input and previously predicted events, so that it can predict the next individual event rather than the entire sequence, thus enhancing controllability. Finally, the predicted event sequence is sent to a downstream model for story generation.

To the best of our knowledge, this is the first attempt to employ an unsupervised graph-based inference approach with a neural advisor as the event planning framework. A range of experiments are conducted to evaluate the performance of our approach, both on the quality of event sequences and their efficacy in aiding story generation. The results demonstrate that our model significantly outperforms all competitive baselines.\footnote{Our code for reproduction is available at \url{https://github.com/tangg555/NGEP-eventplan}.}

\section{Methodology}
The story generation task is formulated as follows: The given input is a sentence acting as the leading context $ C = \{c_1, c_2, ..., c_n\} $ where $c_i$ denotes the $ i $-th token of the leading context, and the output is a multi-sentence story $ S = \{s_1^1, s_2^1, ..., s_1^2 ..., s_n^m\}$, where $ s_j^i $ denotes the $j$-th token of $ i $-th sentence in a story. The task requires the prediction of an event sequence\footnote{We combine events with special tokens, e.g., ``<$s$> needed get <$\mathit{sep}$> ... <$e$>'', where ``<$s$>'',``<$\mathit{sep}$>'', ``<$e$>'' denote the start, separation, and end of planning, respectively. } $ E = \{e_1, e_2, ..., e_m\}$ as a intermediate input, which is generated according to the leading context $C$ and used to generate a story $S$. $ e_i $ denotes the $ i $-th event representing the $ i $-th sentence in a story, and each event may have multiple tokens.

\subsection{Event Graph Construction}

The representation of an event is defined as a verb phrase that describes the main event within a sentence. We employ $\mathit{spaCy}$\footnote{\url{https://spacy.io/}} to parse dependencies between words in a given sentence, and then extract all key roles to compose an event. Neighboring events are considered to have directed relations $r$ (previous/next event), so that each story may contain several triplets $\{e_{\mathrm{head}}, r, e_{\mathrm{tail}}\}$. The set of all triplets in the training set is the event graph $G$. The sum of repeated triplets of an event in the training set is recorded as weighted degrees $d$ in $G$ for calculations of the conditional probability between events. Due to space constraints, the details of the event schema and extraction framework are described in the Appendix (\ref{apx:schema} and \ref{apx:planning}, respectively). 


\subsection{Graph-based Event Planning}

Due to there being no single unique storyline for a given topic, we argue that the planned event sequences for open-domain story generation should instead focus on the intrinsic relatedness between events and their relevance to the leading context. Therefore, we reference the framework of \citet{bamman-smith-2014-unsupervised} and propose an unsupervised graph-based approach to model the conditional probability distribution between events in the event graph $G$. The event contained within the leading context denoted as $e_c$ is set to be the start of the event planning process. Let $P(e_i^\prime|E_{e_{t<i}}^c, G)$ denote the conditional probability of candidates for the $i$-th event $e_i$, and  $E_{e_{t<i}}^c = \{e_c, e_1,... e_{i-1}\}$ denote the input of prior events for the prediction of $e_i$. $P(e_i^\prime|E_{e_{t<i}}^c, G)$ is calculated as follows:

\begin{align}
    P(e_i^\prime|E_{e_{t<i}}^c, G) & = \frac{f_s(r(e_{i-1}, e_{i}^\prime))}{\sum_{r(e_{i-1}, *) \in G} f_s(r)} 
    \label{eq:prob-dist} \\
    f_s(r(e_{i-1}, e_{i}^\prime)) &= \omega(e_{i-1}, e_{i}^\prime) d_{e_{i}^\prime} \times \gamma_(e_{i}^\prime|E_{e_{t<i}}^c) \\
    \gamma_(e_{i}^\prime|E_{e_{t<i}}^c) & = \frac{\left|\mathit{rept}_{\mathit{m}}-c^-(e_{i}^\prime, E_{e_{t<i}}^c)\right|}{\mathit{rept}_{\mathit{m}} \times   d_{*e_{i}^\prime}^{in}}  \\
    e_i & \stackrel{sampling}{\Longleftarrow} P(e_i^\prime|E_{e_{t<i}}^c, G)) 
\end{align}
where $\gamma_(e_{i}^\prime|E_{e_{t<i}}^c)$ denotes the repetition penalty of a candidate $e_{i}^\prime$ ranging from $0$ to $1$, and $\mathit{rept}_{\mathit{m}}$ denotes the maximum number of repetitions permitted in $E_{e_{t<i}}^c$. We penalise candidates with its weighted in-degree $d_{*e_{i}^\prime}^{in}$, as this means it has a relatively weak relationship to $e_{i-1}$. $c^-(e_{i}^\prime, E_{e_{t<i}}^c)$ counts the occurrences of $e_{i}^\prime$ observed in $E_{e_{t<i}}^c$. 

$f_s(r(e_{i-1}, e_{i}^\prime))$ is the event score function which evaluates the probability of event $e_{i}^\prime$ through the calculation of the weight of edge $\omega(e_{i-1}, e_{i}^\prime)$ (as the graph is isomorphic, we set it to 1 here) and the degrees of the event node $d_{e_{i}^\prime}$. Furthermore, $r(e_{\mathit{head}}, e_{\mathit{tail}})$ denotes the directed edge from the head event pointing to the tail event, with $*$ acting as the wildcard character representing any available event. $P(e_i^\prime|E_{e_{t<i}}^c, G)$ is calculated using the event score function and the repetition penalty. Finally, we select the candidate $e_{i}^\prime$ by sampling candidates according to the probability distribution $P$. 

\subsection{Neural Advisor}
\begin{figure}[tb]
\centering
\includegraphics[width=0.9\columnwidth]{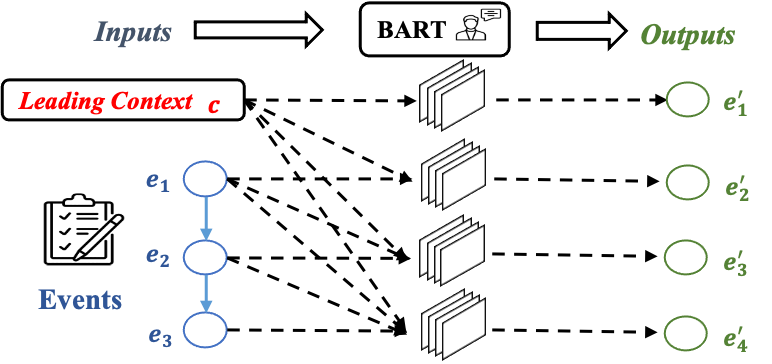}
\caption{Illustration of the neural advisor. }
\label{fig:advisor}
\end{figure}

Event graph inference may not be possible for all instances in the test set if the extracted event from a leading context has not been seen at graph construction time. Consequently, if the event graph is unable to generate any candidates for the next event we need another module to analyse the given information and predict the most probable candidate to compose the storyline. Therefore, as \autoref{fig:advisor} shows, we train a generation model, BART, to "advise" on selecting the next event as below:

\begin{align}
    & E_{e_{t<i}} \{e_1, ..., e_{i-1}\} \quad \mathrm{s.t.} e_t \in G \\
    & F_i = \mathit{Encoder}([C; E_{e_{t<i}}])\\
    & {e_i^{\prime}} \stackrel{predict}{\Longleftarrow} \mathit{Decoder}(F_i)
\end{align}
where $E_{e_{t<i}}$ denotes the prior event sequences before time step $i$. When training, we force BART to learn the relations between reference events, and then find the closest event candidate $e_i^{\prime}$ via the Jaccard similarity index in $G$ to be the next event $e_i$.

\begin{algorithm}[ht]
\DontPrintSemicolon
    \KwInput{A leading context $C$ and the event graph $G$, the minimal planning size of events $ l_{\mathrm{min}} $ and  the maximal $ l_{\mathrm{max}} $}
    \KwOutput{Event Sequence $ E $ for $ C$}
    Initialize $ E \leftarrow $ [];
    
    extract $ e_c $ from $ C $
    
    \If{$ e_c \not \in G $}{
        reselect $e_c \leftarrow e_c^\prime \in G$ where $e_c^\prime$ is equal $e_c.verb$, otherwise $e_c \leftarrow \mathit{advise}(e_c) $        
    }
    
    $e_{\mathrm{pre}} \leftarrow e_c$
    
    \While{$ \left|E\right| < l_{min} $ or $ \left|E\right| > l_{max}$} {
        Let $E^\prime$ denote the set of candidates $e_{\mathrm{next}}^\prime$
        
        \eIf{$ E^\prime = \varnothing $}{
             $e_{\mathrm{next}} \leftarrow \mathit{advise}(e_{\mathrm{next}}) $ 
        }{
            Get $\gamma_(e_{\mathrm{next}}^\prime|E_{e_{t<\mathrm{next}}}^c)$ for $E^\prime$

            Get $P(e_i^\prime|E_{e_{t<i}}^c, G)$ for $E^\prime$
            
            Sample $e_{\mathrm{next}}$ according to $P$ 
                
        }
        Append $e_{\mathrm{next}}$ to $ E $
    }
\caption{Predict Event Sequence $ E $}
\label{alg:prediction}
\end{algorithm}


\subsection{Overall Event Planning Process}
We combine the graph-based event planning with the neural advisor (denoted as $advise(*)$) to predict event sequences (illustrated in \autoref{alg:prediction}). The training objective of neural advisor is same to the vanilla BART, and the graph-based event planning process is unsupervised. 

\section{Experiment}
\subsection{Experiment Setup}
\paragraph{Datasets}
We conduct our experiments on ROCStories \cite{mostafazadeh-etal-2016-corpus}, following the work of \citet{guan-etal-2021-long} to preprocess and split the data. The total number of stories in the Train/Dev/Test sets is 88344/4908/4909.

\paragraph{Training Details and Parameters}
Experiments were performed on an RTX A5000 GPU, and the random seed was fixed to $42$ to facilitate reproduction. We implement the PyTorch Lightning\footnote{\url{https://www.pytorchlightning.ai/}} framework to set up training processes. The training parameters are as follows: \textit{batch size} is set to $64$; \textit{learning rate} is $1\mathrm{e}{-4}$; \textit{max source length} is set to $1024$; the optimiser uses $\mathit{Adam}$ \cite{kingma2014adam}, and the $\epsilon$ of $\mathit{Adam}$ is set to $1\mathrm{e}{-8}$. The whole training process runs for $5$ $\mathit{epochs}$, but the results only consider the checkpoint with the best performance (lowest loss).

\begin{table}[ht]
\centering
\resizebox{1.0\linewidth}{!}{
\begin{tabular}{l|ccc|cc|cc}
\toprule[1pt]
 Metrics & \textbf{R-1}$ \uparrow $ & \textbf{R-2}$ \uparrow $ & \textbf{R-L}$ \uparrow $ & \textbf{B-1}$ \uparrow $ & \textbf{B-2}$ \uparrow $ & \textbf{D-1}$ \uparrow $  & \textbf{D-2}$ \uparrow $  \\
\midrule
\textbf{Seq2Seq}  & 54.33 & 29.10 & 53.05 & 0.391 & 0.089 & 0.051 & 0.277 \\
\textbf{BART}  & 56.36 & 30.35 & 54.68 & 0.398 & 0.095 & \uline{0.060} & 0.298 \\
\textbf{GPT-2}  & 44.78 & 20.71 & 42.80 & 0.217 & 0.052 & 0.055 & \textbf{0.318} \\
\midrule
\textbf{EventAdvisor}  & \textbf{59.85} & \textbf{32.43} & \textbf{57.74} & \textbf{0.436} & \textbf{0.110} & 0.050 & 0.257 \\
\textbf{NGEP}  & \uline{59.30} & \uline{31.96} & \uline{57.54} & \uline{0.429} & \uline{0.099} & \textbf{0.072} & \uline{0.311} \\
\midrule
\textbf{Golden}  & N/A & N/A & N/A & N/A & N/A & 0.072 & 0.315 \\
\bottomrule[1pt]
\end{tabular}
}
\caption{Automatic evaluation on event sequences. $\uparrow/\downarrow$ means the higher/lower the metric, the better. The best performing model is highlighted in \textbf{bold}, and the second best is \uline{underlined}.}
\label{tab:planning}
\end{table}

\begin{table*}[ht]
\centering
\resizebox{0.95\linewidth}{!}{
\begin{tabular}{l|cccc|cccc|cccc|cccc}
\toprule[1pt]
\multirow{2}{*}{} & \multicolumn{4}{c|}{\textbf{Seq2Seq$_{story}$}} & \multicolumn{4}{c|}{\textbf{BART$_{story}$}} & \multicolumn{4}{c|}{\textbf{HINT$_{story}$}} & \multicolumn{4}{c}{\textbf{T-5$_{story}$}} \\
 & \textbf{IR-A$\downarrow$} & \textbf{D-2$\uparrow$}  & \textbf{D-3$\uparrow$} & \textbf{D-4$\uparrow$} & \textbf{IR-A$\downarrow$} & \textbf{D-2$\uparrow$}  & \textbf{D-3$\uparrow$} & \textbf{D-4$\uparrow$} & \textbf{IR-A$\downarrow$} & \textbf{D-2$\uparrow$}  & \textbf{D-3$\uparrow$} & \textbf{D-4$\uparrow$} & \textbf{IR-A$\downarrow$} & \textbf{D-2$\uparrow$}  & \textbf{D-3$\uparrow$} & \textbf{D-4$\uparrow$} \\
\midrule
\textbf{w/o events}  &  \textbf{1.16} & 0.233 & 0.554 & 0.777 & 1.88 & 0.243 &  0.567 & 0.789 & 1.81 &  \uline{0.188} & \uline{0.494} & \uline{0.740} & 1.68 &  0.216 & \uline{0.498} & \uline{0.719}  \\
\midrule
\textbf{Seq2Seq}  &  1.27 & 0.227 & 0.546 & 0.773 & \uline{1.40} & 0.247 &  \uline{0.576} & \uline{0.799} & \uline{1.43} &  0.185 & 0.490 & 0.738 & \uline{1.54} &  0.213 & 0.497 & \uline{0.719} \\
\textbf{BART} &  1.33  & 0.230 & 0.547 & 0.769 & 1.74 & \uline{0.250} & 0.575 & 0.795 & 1.76 &  \uline{0.188} & 0.490 & 0.732  & 1.93 &  \uline{0.218} & \uline{0.498} & \uline{0.719} \\
\textbf{GPT-2}  & \uline{1.25} & 0.222 & 0.544 &  0.776 & 1.98 & 0.235 &  0.565 & 0.791 & 1.87 &  0.174 & 0.472 & 0.720 & 2.32 &  0.209 & 0.493 & 0.718  \\
\midrule
\textbf{EventAdvisor}  &  1.32 & \uline{0.234} & \uline{0.555}  & \uline{0.778} & 1.75 & 0.244 & 0.564 & 0.781  & 1.80 &  0.183 & 0.478 & 0.718 & 1.84 &  0.211 & 0.490 & 0.712 \\
\textbf{NGEP}  &  \textbf{1.16} &  \textbf{0.235} & \textbf{0.558} & \textbf{0.779}   & \textbf{1.31} & \textbf{0.272} &  \textbf{0.601} & \textbf{0.811} & \textbf{1.25} &  \textbf{0.244} & \textbf{0.507} & \textbf{0.742} & \textbf{1.29} &  \textbf{0.231} & \textbf{0.517} & \textbf{0.738} \\
\bottomrule[1pt]
\end{tabular}
}
\caption{Automatic evaluation with unreferenced metrics on generated stories. The row labels stand for different event planning methods, and the column labels are SOTA models for story generation. }
\label{tab:story}
\end{table*}

\begin{figure*}[ht]
\centering
\includegraphics[width=1.9\columnwidth]{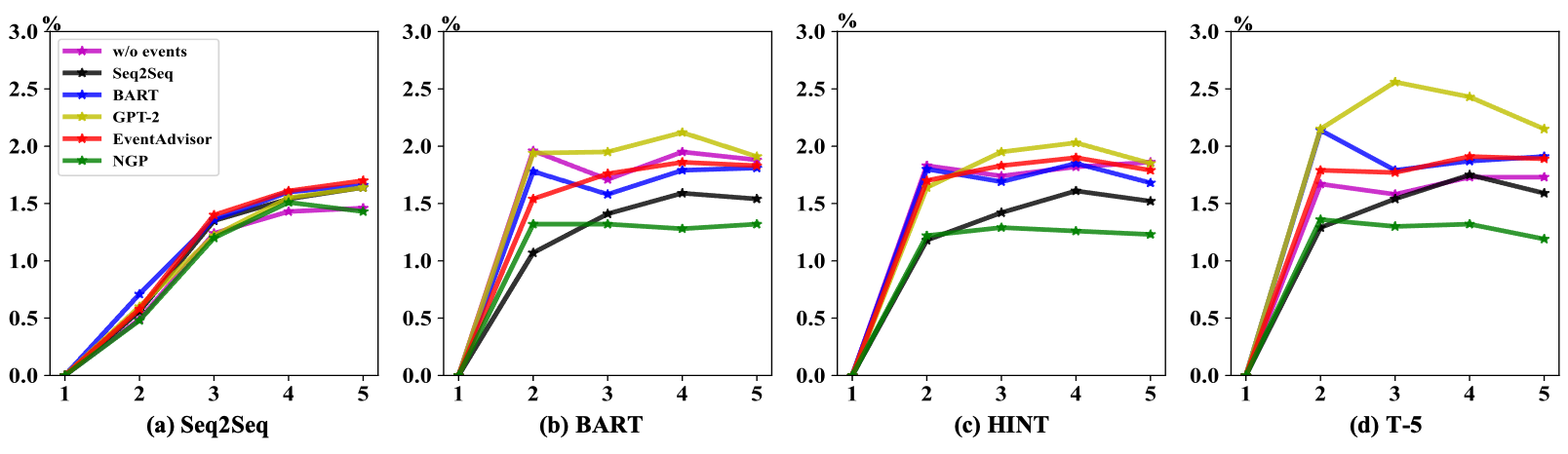}
\caption{Intra-story repetitions (the lower the better) for each sentence in a story. We show the performance of different event planning approaches work different story generation models.}
\label{fig:repetition}
\end{figure*}

\paragraph{Baselines}
Several SOTA generation models for event planning and story generation (or long text generation) are selected as baselines.\footnote{We additionally intended to compare our model to GraphPlan \cite{chen-etal-2021-graphplan}, which also proposed the use of event graphs to improve event planning. However, we encountered difficulties in attempting to reproduce this work, e.g., the word embedding based framework only works for one-word events and there is no publicly available code.} (\romannumeral1) \textbf{Neural Event Planning:} Seq2Seq \cite{yao2019plan}, BART \cite{goldfarb-tarrant-etal-2020-content}, and GPT-2 \cite{chen-etal-2021-graphplan}; (\romannumeral2) \textbf{Story Generation} Seq2Seq \cite{yao2019plan}, BART \cite{goldfarb-tarrant-etal-2020-content}, HINT \cite{guan-etal-2021-long}, and T-5 \cite{raffel2020exploring}, in line with previous work in the area.

\subsection{Evaluation Metrics}
We adopt a range of automatic metrics including \textbf{ROUGE-n (R-n)} \cite{lin-2004-rouge} and \textbf{BLEU-n (B-n)} \cite{papineni-etal-2002-bleu} as referenced metrics to compare to human-written event plans, and
\textbf{Distinction-n (D-n)} \cite{li-etal-2016-diversity}, \textbf{Intra-story Repetition} \citep{yao2019plan}, and \textbf{Intra-story Repetition Aggregate Score (IR-A)} \citep{yao2019plan} to assess the degree of repetition and diversity within event sequences and generated stories.

\subsection{Experimental Results}
\paragraph{Evaluation of Event Sequences}

As shown in \autoref{tab:planning}, when considering all metrics, both EventAdvisor and NGEP substantially outperform the selected baselines. Performance on the referenced metrics, $\mathit{ROUGE}$ and $\mathit{BLEU}$, indicates that the events predicted by our proposed models are more similar to the human-written event sequences. We hypothesise that the superior performance of EventAdvisor over NGEP is a result of select test events not being present in $G$, with our event advisor being more robust to such cases.

\paragraph{Performance on Story Generation} 
\autoref{tab:story} measures the quality of generated stories\footnote{$C$ and $E$ are concatenated as the input of those models.} on unreferenced metrics conditioning on the leading context $C$ and event plans $E$. We observe that NGEP substantially outperforms all baseline models. This indicates that our proposed graph-based inference improves story generation through planning better storylines, as our predicted events have no hallucination problems and contain event sequences that are more logically coherent. The intra-story repetitions shown in \autoref{fig:repetition} further demonstrate that the proposed model is more stable throughout the generation process (less fluctuations), and the predicted events display less repetition, improving the diversity of stories.

\begin{figure}[ht]
\centering
\includegraphics[width=1.0\columnwidth]{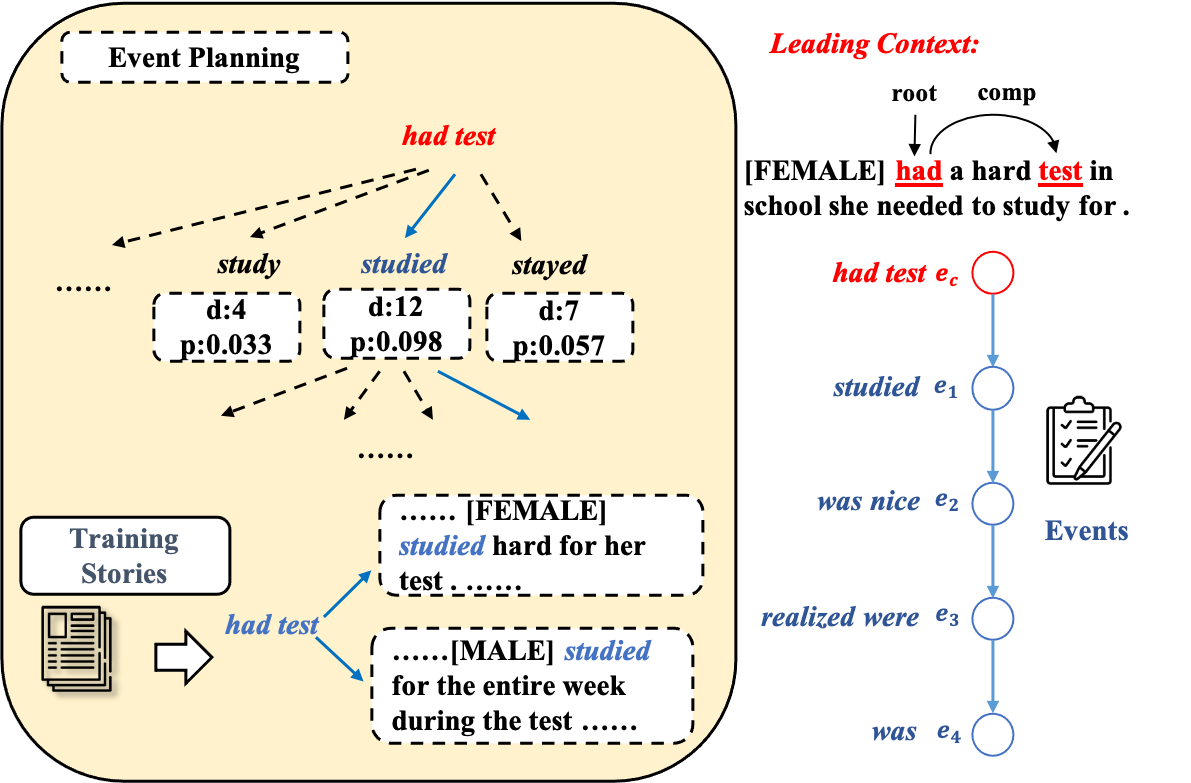}
\caption{An example of the event planning process within our proposed NGEP. $d$ denotes degree, and $p$ denotes the conditional probability.}
\label{fig:case}
\end{figure}

\paragraph{In-depth Analysis} 
\label{sec:quantity}
To further study how the proposed framework works during event planning, we conduct a case study as illustrated in \autoref{fig:case}. Given the leading context, we can extract the contained event \textbf{\textit{had test}}. In the event graph constructed from the training dataset, the event \textbf{\textit{had test}} has many candidates whose conditional probabilities are calculated by the proposed NGEP. It can be observed that the event candidate \textbf{\textit{studied}} has the highest probability. This is because, in the training dataset, more stories contain the content "people studied hard to prepare for this test". This indicates that instead of implicitly capturing the relatedness between events through neural models, NGEP allows the predicted events to have more knowledge grounding. Therefore, compared to traditional neural event planning methods, the processes behind NGEP are easier to interpret, whilst also avoiding the hallucination problem of deep learning.

\section{Conclusion}
This study proposes a novel hybrid event planning approach which performs inference on event graphs with the help of a neural event advisor. A range of experiments demonstrate that the proposed model outperforms other SOTA neural event planning approaches, and substantially improves performance on the downstream task of story generation.

\section*{Acknowledgements}
Chen Tang is supported by the China Scholarship Council (CSC) for his doctoral study (File No.202006120039). Tyler Loakman is supported by the Centre for Doctoral Training in Speech and Language Technologies (SLT) and their Applications funded by UK Research and Innovation [grant number EP/S023062/1]. We also gratefully acknowledge the anonymous reviewers for their insightful comments.

\normalem
\bibliography{bibs/sec1-introduction,
              bibs/sec2-methodology,
              bibs/sec3-experiment}

\appendix
\section{Appendix}
\label{sec:appendix}

\subsection{Details of Event Schema} \label{apx:schema}

\begin{figure}[htpb]
\centering
\includegraphics[width=\columnwidth]{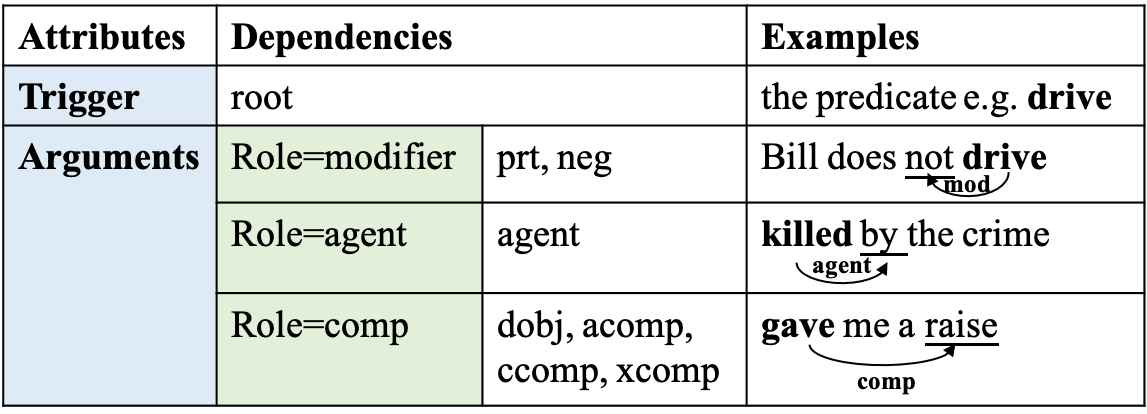}
\caption{The schema of event shows the relations with event arguments and word dependencies. We offer some examples to indicate these dependencies, e.g., in "Bill does not drive", "not" is a negation (\textbf{neg}) of "drive", so it is an event modifier. }
\label{fig:schema}
\end{figure}

An event is intended to represent an important change that happens within a narrative, and so generally represents an action. The schema for an event aims to include all relevant roles to the action (e.g., verbs and object) and filter trivial details for representation. Inspired by the work of \citet{rusu-etal-2014-unsupervised} and \citet{bjorne-salakoski-2018-biomedical} which used dependency parsing to capture dependencies between words belonging to different clauses, we extract event mentions from sentences according to the hierarchy of typed dependencies \citep{de2008stanford} (see details in Appendix.~\ref{apx:schema}). In this way we can obtain more informative and unambiguous events compared to single-verb representations used in previous work \cite{jhamtani-berg-kirkpatrick-2020-narrative, guan-etal-2020-knowledge}. The schema is shown in Figure \ref{fig:schema}. 

As shown in Figure.~\ref{fig:schema}, event arguments are extracted according to selected dependencies between words. Here, we give the details of these dependencies, and Table.~\ref{tab:schema_details} indicates the roles of these dependencies in a sentence (for more details of dependencies see \citet{de2008stanford}).
\begin{table}[htpb]
\centering
\resizebox{\linewidth}{!}{
\begin{tabular}{lcc}
\toprule[1pt]
\textbf{Dep.} & \textbf{Full Name} & \textbf{Example} \\
\hline

 \textbf{prt} & phrasal verb particle & [shut]-\textit{prt}->[down] \\
 \textbf{neg} & negation modifier & [drive]-\textit{neg}->[not] \\
 \textbf{agent} & agent & [killed]-\textit{agent}->[police] \\
 \textbf{dobj} & direct object & [gave]-\textit{dobj}->[raise] \\
 \textbf{acomp} & adjectival complement & [looks]-\textit{acomp}->[beautiful] \\
 \textbf{ccomp} & clausal complement & [says]-\textit{comp}->[like] \\
 \textbf{xcomp} & open clausal complement & [like]-\textit{xcomp}->[swim] \\
 
\bottomrule[1pt]
\end{tabular}
}
\caption{Details of dependencies in Event Schema. Examples are extracted with the format [head]-\textit{dependency}->[tail].}
\label{tab:schema_details}
\end{table}

The schemas of events are required to consider performance with respect to both generalisation and representation. The more dependencies included, the more potentially informative an event may become, at the cost of reduced generalisation. For instance, the \textit{Subject} (e.g. I, you, Kent, etc.) is useful to identify the protagonist of an event, but stories usually have different characters, making it challenging to reuse events from one story in another. For example, "Kent is driving" and "He is driving" refer to the same semantic event, but if "Kent" is extracted as an event unit, it is very hard to predict the same event for another story, which means generalisation is impaired. According to a similar criterion, we select key roles as the arguments of events with the consideration of both generalisation and representation.

\subsection{Details of Event Extraction} \label{apx:planning}
We extract events from the text of the training dataset including reference stories and leading contexts. The data structure of an event is a set including the relevant triggers and arguments in a sentence. We firstly use $\mathit{spaCy}$ to parse dependencies between words in a sentence, and then annotate the event trigger and arguments according to their dependencies. An event  $ e $ contains attributes introduced in Figure \ref{fig:schema}, in which the event trigger is usually the predicate. Before encoders accept text as the input, the extracted events are serialised to text format to pass to the model.

Since existing story datasets do not have the reference storylines paired with reference stories, we develop an event extractor that extracts event sequences from reference stories to act as the storylines. We follow the approach of representing events as verb phrases. Verbs, as the anchor of sentences, can be seen as the \textit{event trigger}, so our primary goal is to extract all key roles (as \textit{event arguments}) related to the event trigger. The neighbourhood of extracted events will be considered as temporal relations. 

With the temporally related events from the training stories, we construct an event graph denoted $ G $, which is an isomorphic graph with a single event type and a single relation type. We suppose $ G $ is a data structure composed of triples in $e_h, r, e_t$ format. The workflow of the extraction process is explained as follows:

\begin{algorithm}[!ht]
\DontPrintSemicolon
  \KwInput{A story $ S $ with $ m $ sentences}
  \KwOutput{Event Sequence $ E $ for $ S $  containing $ m $ event objects}
  Initialise $ E \leftarrow \varnothing $ and $ roles \leftarrow \{\mathit{trigger}, \mathit{mod}, \mathit{agent}, \mathit{comp}\} $
  \ForEach{$s^i$ in $S$}{
        Initialise $ e_i \leftarrow \varnothing $
        
        Normalise $ s^i $ and get dependencies $\mathit{dep}_i$ with $\mathit{spaCy}$
        
        Extract event trigger $t$ and position $p_t$ from $\mathit{dep}_i$
        
        $ e_i.\mathit{trigger} \leftarrow t$
        
          \ForEach{$\mathit{role}$ in $\mathit{role}$}{
              \If{$t \in \mathit{dep}_i.heads$ and $\mathit{role} \in \mathit{dep}_i.tails $}{
                  Extract $(\mathit{role}, p_{\mathit{r}})$ from $\mathit{dep}_i$
            
                  $ e_i.\mathit{role} \leftarrow (\mathit{role}, p_{\mathit{r}}) $
              }
      }
      $ e_i.\mathit{string} \leftarrow r \in \mathit{roles} $ aligned by $ p_r \uparrow$
      
      $E$ append $e_i$
  }
\caption{Extract Event Sequence $E$\label{alg:extract}}
\end{algorithm}

\end{document}